\documentclass[conference]{IEEEtran}
\IEEEoverridecommandlockouts
\usepackage{cite}
\usepackage{amsmath,amssymb,amsfonts}
\usepackage{algorithmic}
\usepackage{graphicx}
\usepackage{textcomp}
\usepackage{soul}
\usepackage{xcolor}

\newcommand{\mycomment}[1]{}

\usepackage{tikz}
\usepackage{hyperref}
\usepackage{lipsum}

\newcommand\copyrighttext{%
  \footnotesize \textcopyright 2025 IEEE. Personal use of this material is permitted.
  Permission from IEEE must be obtained for all other uses, in any current or future 
  media, including reprinting/republishing this material for advertising or promotional 
  purposes, creating new collective works, for resale or redistribution to servers or 
  lists, or reuse of any copyrighted component of this work in other works. The definitive version was published in IEEE International Smart Cities Conference (ISC2) Prooceedings, DOI:  \href{https://doi.org/10.1109/ISC266238.2025.11293252}{10.1109/ISC266238.2025.11293252}.
  }
\newcommand\copyrightnotice{%
\begin{tikzpicture}[remember picture,overlay]
\node[anchor=south,yshift=10pt] at (current page.south) {\fbox{\parbox{\dimexpr\textwidth-\fboxsep-\fboxrule\relax}{\copyrighttext}}};
\end{tikzpicture}%
}

\def\BibTeX{{\rm B\kern-.05em{\sc i\kern-.025em b}\kern-.08em
    T\kern-.1667em\lower.7ex\hbox{E}\kern-.125emX}}
\begin{document}

\title{Digital Twin based Automatic Reconfiguration of Robotic Systems in Smart Environments\\
\thanks{This work was partially funded by the National Recovery and Resilience Plan Greece 2.0, funded by the European Union – NextGeneration EU, Greece4.0 project, under agreement no. TAEDR-0535864. The paper reflects the authors’ views, and the Commission is not responsible for any use that may be made of the information it contains.}
}

\author{
    \IEEEauthorblockN{Angelos Alexopoulos\IEEEauthorrefmark{1}\IEEEauthorrefmark{2},
    Agorakis Bompotas\IEEEauthorrefmark{1},
    Nikitas Rigas Kalogeropoulos\IEEEauthorrefmark{1},\\
    Panagiotis Kechagias\IEEEauthorrefmark{1},
    Athanasios P. Kalogeras\IEEEauthorrefmark{1},
    Christos Alexakos\IEEEauthorrefmark{1}}
    \\
    \IEEEauthorblockA{\IEEEauthorrefmark{1}\textit{Industrial Systems Institute} \\
    \textit{ATHENA Research Center}\\
    Patras, Greece \\
    \{aggalexopoulos, abompotas, nkalogeropoulos, kechagias, kalogeras, alexakos\}@athenarc.gr}
    \\
    \IEEEauthorblockA{\IEEEauthorrefmark{2}\textit{Physics Department} \\
    \textit{University of Patras}\\
    Patras, Greece }

%

}

\maketitle

\copyrightnotice

\begin{abstract}
Robotic systems have become integral to smart environments, enabling applications ranging from urban surveillance and automated agriculture to industrial automation. However, their effective operation in dynamic settings—such as smart cities and precision farming—is challenged by continuously evolving topographies and environmental conditions. Traditional control systems often struggle to adapt quickly, leading to inefficiencies or operational failures. To address this limitation, we propose a novel framework for autonomous and dynamic reconfiguration of robotic controllers using Digital Twin technology. Our approach leverages a virtual replica of the robot’s operational environment to simulate and optimize movement trajectories in response to real-world changes. By recalculating paths and control parameters in the Digital Twin and deploying the updated code to the physical robot, our method ensures rapid and reliable adaptation without manual intervention. This work advances the integration of Digital Twins in robotics, offering a scalable solution for enhancing autonomy in smart, dynamic environments.

\end{abstract}


\section{Introduction}
Robotic systems have achieved significant penetration in smart environments, including smart cities, smart manufacturing, and smart agriculture. Smart unmanned vehicles that deliver services inside cities, drones that monitor critical infrastructure of the city (i.e. traffic), UAVs that assist tourists, or even robotics that help pedestrians and especially persons with mobility impairments, exemplify the integration of robotics into smart city ecosystems ~\cite{tiddi2020robot}. Moreover, robots have become an integral part of modern agricultural infrastructure, automating tasks and assisting workers in their daily operations ~\cite{pal2023ai}. All these robotic systems require precise digital control, typically provided by industrial-grade controllers similar to those used in other smart infrastructures like buildings and factories. Another characteristic of smart environments is their changeability over time, especially regarding the choreographic aspect. It is common for a city to construct new buildings, roads or parks causing changes in the landscape, while this also stands true in the agricultural sector with changes in crop fields as well as relevant processing and storage infrastructures \cite{el2021smart}. Changes in the topology of smart environments demand quick and dependable controller reconfiguration,  so that they can adapt to changes and effectively plan and execute robotic system movements. 

The concept of Digital Twins (DTs) has gained significant traction in the robotics domain, offering real-time synchronization between physical assets and their virtual counterparts ~\cite{mylonas2021digital}. Early applications focused on offline simulation and predictive maintenance in industrial robotics~\cite{tao2018digital}, but recent work emphasizes their role in runtime monitoring, control, and autonomy. For instance, Lu et al.~\cite{lu2020digital} proposed a cloud-based DT framework for collaborative robot management, enabling status tracking and fault detection. Similarly, Schleich et al.~\cite{schleich2017shaping} introduced hybrid physical-virtual models for robot diagnostics and performance prediction. Despite these advancements, most DT implementations in robotics remain static or limited to visualization, with few addressing closed-loop feedback for system reconfiguration.


The present work introduces a novel approach to seamlessly integrate real and virtual environments in Digital Twin implementations by closing the feedback loop. The user can provide the DT with new information about the monitored real physical environment regarding modifications to the scene, including introduction of new objects or removal of existing ones. 
The DT updates the robotic system's motion plan according to the modified environment and sends the revised control commands back to the physical setup. 
This approach enhances DTs with flexibility and scalability, enabling them to adapt to constantly changing object arrangements in real-world smart environments.

The rest of the paper is structured as follows. Section II presents the background and related work on the utilization of DTs in smart environments, robotic movement trajectory calculation and robotic controller reconfiguration, section III details the proposed DT-driven approach, while section IV presents a relevant use-case. Finally, section V provides discussion and conclusions.

\section{Background \& Related Work}

\subsection{Robotic Simulation Platforms and Digital Twin Environments}

Simulation platforms play a crucial role in the development and validation of DTs for robotic systems. A recent empirical study by Singh et al.~\cite{singh2025comparative} offers a detailed comparison of two prominent environments—Unity3D and Gazebo—through the development of a DT of an ABB IRB 1200 robotic arm. This investigation highlights both platforms’ strengths and limitations across metrics such as accuracy, latency, graphics rendering, ROS integration, cost, and scalability~\cite{singh2025comparative}.

Unity stands out for its high-fidelity real-time rendering, enabling visually immersive DTs with accurate trajectory replication. In the study, Unity produced significantly lower latency $(77.7 \pm 15.7ms)$ and superior graphical realism compared to Gazebo~\cite{singh2025comparative}. However, Unity’s integration with ROS requires intermediate tools (e.g., ROS\# or ROS–TCP–Connector), which introduces complexity and a steeper learning curve~\cite{singh2025comparative}. Despite this, Unity supports cross-platform deployment, including AR/VR devices, and benefits from fast prototyping workflows and robust asset libraries.

On the other hand, Gazebo excels in native ROS compatibility and accurate physics simulation. The study reports latency of $108.8 \pm 57.6ms$ and somewhat lower graphic fidelity in comparison to Unity~\cite{singh2025comparative}, but emphasizes Gazebo’s no-cost, open-source nature—making it suitable for research and educational projects. Its deep sensor emulation, modular plugins, and standardized URDF-based simulations simplify DT implementation without substantial integration overhead~\cite{singh2025comparative}.

The comparative findings suggest that while Unity excels in visualization and low-latency interaction, Gazebo remains advantageous in scenarios where accurate physics simulation and ROS-native behavior are paramount. Hybrid approaches that leverage the strengths of both platforms—such as combining Gazebo’s physics engine with Unity’s rendering through frameworks like OpenUAV~\cite{anand2021openuav}—offer a promising direction for more robust DT implementations.

Moreover, Unity's ML-Agents toolkit~\cite{juliani2018unity} further extends its potential for AI-driven control and reinforcement learning, which is increasingly relevant for autonomous robotic systems. Nevertheless, Singh et al.~\cite{singh2025comparative} emphasize that neither Unity nor Gazebo has yet been fully exploited for enabling runtime reconfiguration of robots through DTs in dynamic smart environments. This underscores a current gap in simulation-driven reconfigurable robotics, motivating the need for new frameworks that integrate high-fidelity DTs with autonomous system adaptation.
Finally, an important distinction between Gazebo and game engines like Unity, is the latter's support for deformable mesh geometries \cite{senthilkumar2023simulating}, as game engines usually feature extended support for meshes, editors, importers and the like, while simpler simulation environments require extensive work on incorporating addons, and even then real-time mesh alteration and deformation is not guaranteed.

\subsection{Trajectory Planning in Smart Robotic Systems}

Trajectory planning is a fundamental component of autonomous robotic behavior, particularly within smart environments that demand continuous adaptation. In the context of DTs, trajectory planning serves as a critical bridge between virtual simulation and real-world execution. The literature typically segments this process into three core phases:
\begin{enumerate}
    \item environment modeling
    \item path planning
    \item trajectory execution
\end{enumerate}

The first phase involves the generation of a 3D representation of the robot's surroundings. This is commonly achieved using Simultaneous Localization and Mapping (SLAM) algorithms, which construct spatial maps based on onboard sensor data~\cite{cadena2016past}. Techniques such as GMapping, Cartographer, and RTAB-Map are widely adopted for both indoor and outdoor scenarios~\cite{labbe2019rtab}. In simulated environments, tools such as Unity3D and Gazebo are used to emulate SLAM pipelines~\cite{singh2025comparative}, providing safe, repeatable conditions for evaluating localization accuracy and environmental complexity.

The second phase pertains to path planning, which focuses on computing a collision-free path from the robot’s current position to its goal. Classical graph-based algorithms such as A*, Dijkstra’s algorithm, and their dynamic variants like D* and D*-Lite are frequently used for global planning in known or partially known maps~\cite{koenig2002d}. These algorithms are extensively supported in the Robot Operating System (ROS) ~\cite{chitta2012moveit} through packages such as \texttt{nav\_core}, \texttt{global\_planner}, and \texttt{move\_base}. 

In contrast to graph-based techniques, sampling-based methods such as Rapidly-exploring Random Trees (RRT), RRT*, and Probabilistic Roadmaps (PRM) have gained prominence for motion planning in high-dimensional or continuous configuration spaces~\cite{lavalle1998rapidly,kavraki1996probabilistic,karaman2011sampling}. These methods do not require an explicit discretization of the workspace; instead, they construct a roadmap or tree by randomly sampling feasible states and connecting them through local planners. Their efficiency and scalability make them especially suitable for complex robotic systems with many degrees of freedom, such as manipulators or mobile manipulators. In the context of ROS, these planners are integrated through frameworks like \texttt{OMPL} (Open Motion Planning Library), which interfaces seamlessly with \texttt{MoveIt!} for planning in manipulation tasks~\cite{sucan2012open}.

The final phase—trajectory execution—translates the planned path into a sequence of velocity and position commands, constrained by the robot's physical structure, actuator capabilities, and real-time state. A key tool widely used in this phase is the \texttt{MoveIt!} framework~\cite{chitta2012moveit}, which integrates motion planning, kinematics, collision avoidance, and controller management. \texttt{MoveIt!} interfaces with ROS and simulation tools like Gazebo to execute trajectories that account for real-world dynamics and safety constraints. Its use of sampling-based planners from OMPL enables support for a wide range of robotic configurations, from manipulators to mobile platforms.

While these planning layers are well-supported, there is a lack of frameworks that integrate trajectory planning with DTs for closed-loop adaptive reconfiguration in real-time. The present paper addresses this gap by embedding DT feedback into each trajectory planning phase to support robust, self-adaptive robotic behavior in smart environments.

\subsection{Robotics Reconfiguration}

Reconfiguration refers to the system’s ability to adapt its behavior in response to changes in environment, task, or internal status. 
While reconfiguration has been extensively studied in manufacturing ~\cite{Khan2022}, agriculture ~\cite{HERNANDEZ2025110161}, and other systems, advances in complementary technologies—such as AI and communication infrastructure—now offer new perspectives to revisit and enhance this concept.

DT is such a technology, greatly complementing reconfiguration. It presents an opportunity to address the challenges of reconfigurable systems by creating a simultaneous digital environment in which the various configurations of the system can be tested, before being applied in the real-world\cite{grieves2017digital}. Through DTs, it becomes feasible to simulate behavioral changes before execution, assess system-level implications, and guide reconfiguration with predictive feedback~\cite{uhlemann2017digital}. However, most DT implementations in robotics focus on monitoring or visualization tasks, rather than being actively integrated into the decision-making loop~\cite{lu2020digital}.

This disconnect highlights a critical research gap: current frameworks fail to fully leverage Digital Twins (DTs) for both real-time state tracking and autonomous, adaptive reconfiguration. Closing this gap is vital for robotics in smart environments, where systems must dynamically adapt to changing goals, infrastructure constraints, and multi-agent coordination demands ~\cite{lee2016introduction}.

\section{Digital Twin Driven Approach}

The proposed DT-driven approach for robotic system reconfiguration is based on a key principle: the DT must simulate the robot's environment as realistically as possible. While full realism isn’t always required and is potentially not in the scope of DT application, the DT should enable high-fidelity simulation with minimal developer effort ~\cite{rundel2023leveraging},~\cite{clausen2022can},~\cite{konig2025implementation}. This has led to the adoption of powerful game engines for DT development. Additionally, robotics control platforms like ROS are commonly used to expose integration capabilities and companion packages for seamless connectivity between the game engine and robotic systems.

DT's most important dimension lies in its real-time synchronization and exchange of data with its Physical counterpart, as well as the facilitation of its data driven decision making performed in the DT virtual environment ~\cite{es2024methods}. 
For robotic reconfiguration, the Digital Twin simulates key elements—such as mesh geometries, moving objects, and lighting conditions—to compute optimal trajectories, which are then deployed to the physical robot for execution. There is constant exchange of information across the channels connecting the physical and the digital dimensions of the application, with the one complementing the other. 

\subsection{Setting up of the DT}

The proposed DT is based on the utilization of a specialized domain-specific language (DSL), which serves as a high-level configuration and modeling tool for the DT ~\cite{lehner2025model}. This DSL provides a structured and expressive way to define all critical components, relationships, and behaviors of the physical system. It includes metadata and hierarchical descriptions of machines, controllers, communication interfaces, physical layouts, and process logic. The language must be expressive enough to articulate both static and dynamic aspects of the system: from the position, orientation, and dimensions of machines within a 3D spatial frame to the logical interconnections between control devices and machines. In the context of the present work, Automation ML ~\cite{faltinski2012automationml} is used to automatically configure the DT platform and create the 3D scenery in a Unity virtual environment  ~\cite{alexakos2020iot} .

Each machine or physical element in the system is instantiated with a set of parameters defining  its identity, its role but also its spatial constraints and operational limits. These instances are often linked to digital models imported from CAD or mesh files (e.g., COLLADA or STL), allowing for accurate geometry-based simulations and collision checking in the virtual environment. Sensors and actuators are defined with associated characteristics such as sampling rate, signal type (digital/analog) and interface protocol (Modbus, OPC-UA, ROS, MQTT).

Controllers—whether they are Programmable Logic Controllers (PLCs), microcontrollers, or C++/Python-based control scripts—are explicitly described in the DSL and mapped to specific hardware or virtual execution containers. The descriptive language allows users to specify signal routing, event triggers, and command-response behavior, ensuring a faithful mapping between digital and physical layers. Furthermore, logical connections are made between input/output (I/O) signals and processing units, enabling sensor data to be interpreted in real time and used for control decisions, diagnostics, or predictive analytics.

Based on this information, a DT generator software sets up, apart from the 3D scenery, the DT's configuration of communication protocols and middleware layers that support real-time data exchange between physical components and their digital counterparts  ~\cite{alexakos2020iot}. This includes timing constraints and synchronization mechanisms that help maintain coherence between simulation and reality.

\subsection{Calculating Trajectories}

\begin{figure}[ht]
    \includegraphics[width=\columnwidth]{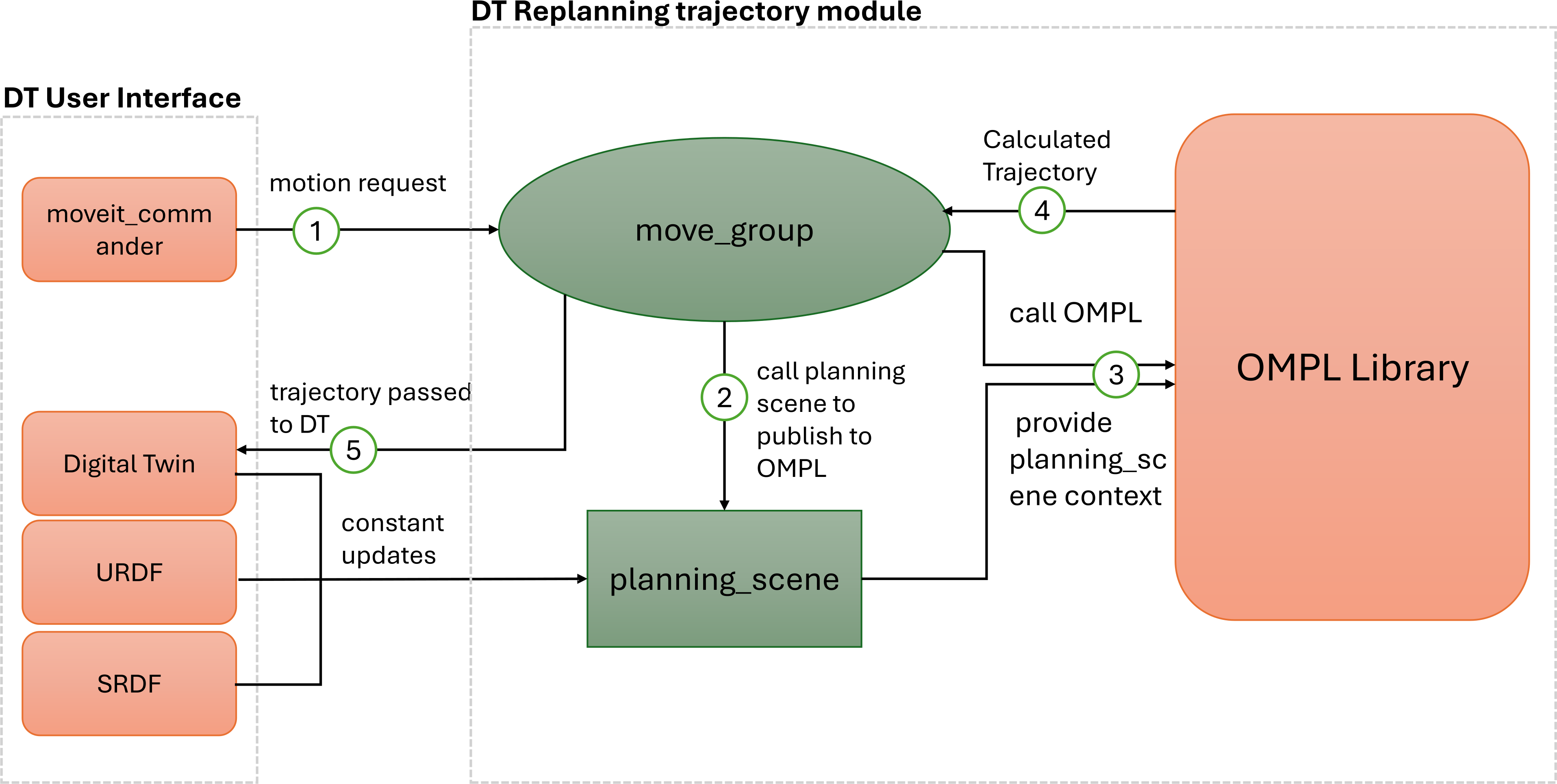}
    \caption{Trajectory planning on Digital Twin}
    \label{fig:moveit_internal}
\end{figure}

In robotic systems using ROS and MoveIt!  for motion planning, the overall process is centrally coordinated by the \texttt{move\_group} node. The integration of a DT into this workflow significantly enhances the system's dynamic understanding of the environment, for accurate and real-time planning. The process, as depicted in Fig. \ref{fig:moveit_internal}, utilizes the OMPL library to plan the new trajectory following user demand and lead to adaptation of the robot controller in the virtual environment accordingly. 

The motion planning process begins when a user defines a motion goal—such as “move the end-effector to this pose”—through high-level APIs like \texttt{moveit\_commander} (Python) or \texttt{MoveGroupInterface} (C++). This request is then forwarded to the \texttt{move\_group} node via ROS services or action interfaces such as \texttt{/move\_group/goal} (Step 1 in Fig.  \ref{fig:moveit_internal}). The \texttt{move\_group} node then takes charge of orchestrating the entire planning pipeline.

Once the goal is received, \texttt{move\_group} queries the \texttt{planning\_scene} to construct the planning context. The \texttt{planning\_scene} is a rich data structure that includes the robot's joint states, frame transforms, known obstacles, joint limits, the allowed collision matrix (ACM), attached objects, and any path or goal constraints (Step 2 in Fig. \ref{fig:moveit_internal}). The DT pushes continuous updates to the \texttt{planning\_scene}, ensuring the environment model reflects changes such as moving obstacles, reconfigured workspaces, or the presence of humans in real time.

With this up-to-date context in hand, \texttt{move\_group} invokes the appropriate motion planning plugin, OMPL, which receives the planning scene and goal as inputs (Step 3 in Fig. \ref{fig:moveit_internal}). Although OMPL itself is unaware of ROS-specific semantics, it works within the configuration space defined by the robot’s kinematics and constraints.

OMPL then begins sampling candidate paths in configuration space using algorithms such as RRT or PRM. For each sample, it invokes MoveIt!'s state validity checkers and motion validators, which in turn query the \texttt{planning\_scene} to confirm that configurations are collision-free and satisfy all constraints. The final result is a trajectory that is returned to  \texttt{move\_group} (Step 4 in Fig. \ref{fig:moveit_internal}) for optional post-processing, which may include smoothing, interpolation, and time parameterization. 

In the final stage, the computed trajectory is returned to the client application. The trajectory is transmitted to the DT for visualization and validation (Step 5 in Fig. \ref{fig:moveit_internal}). 

\begin{figure}[ht]
    \includegraphics[width=\columnwidth]{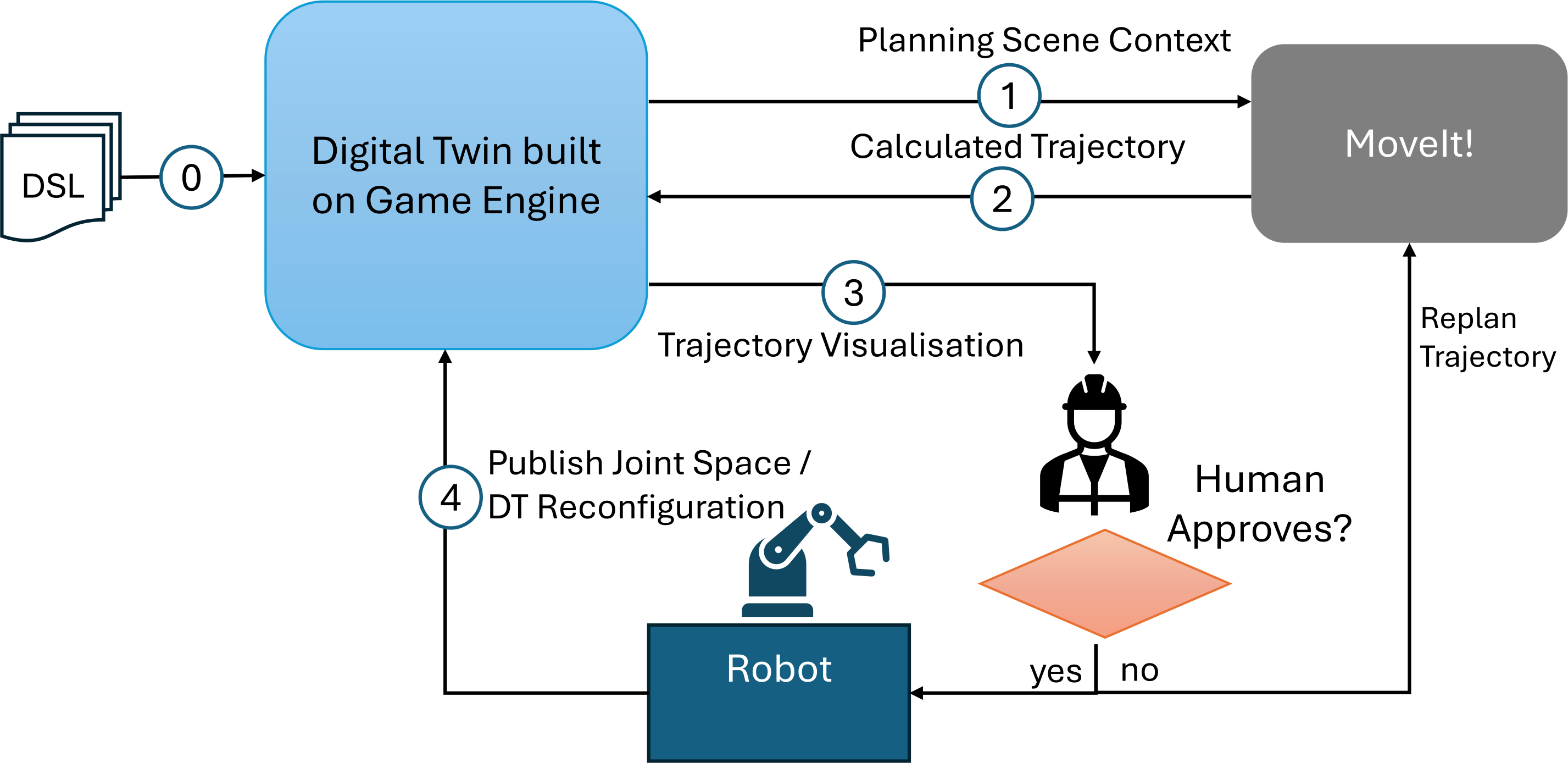}
    \caption{Reconfiguration cycle}
    \label{fig:reconf}
\end{figure}

\subsection{Reconfiguring}
The previously described process involving MoveIt! and the DT paves the way towards the real-time reconfiguration of the robot's state and environment. In the previous section MoveIt!'s internal workings were analyzed in order to contextualize its use in the wider architecture that includes the real Robot, the DT, MoveIt! and a human operator. 

The present sub-section describes the cycle that serves both the digital and physical counterparts of the robot, as depicted in Fig. \ref{fig:reconf}. The initial Step 0 includes the automatic configuration of the DT based on the DSL description. As mentioned in sub-section III.A, this initialization accurately represents all relevant information for the robot (joint space, environmental state). This information is passed to the \texttt{planning\_scene} of MoveIt! during  Step 1 of the reconfiguration cycle, when the new trajectory is calculated. 

The new trajectory information is furnished back into the DT (Step 2), enabling it to adjust the robot's control routines in real time accordingly. Subsequently, a visualized simulation of the new robotic movement enables a human operator to review, intervene and make precise adjustments (Step 3). The operator can take the decision to approve the generated trajectory, or reject it. In the latter case the system attempts to re-plan the trajectory, following a different path and then prompts the operator again for approval.

In the case of human approval, the system proceeds with the transfer of the trajectory control code to the physical robot for execution. While the trajectory is executed, the joint space of the robot, and any of its interactions with the environment are sent back to the DT, reconfiguring the virtual environment in real-time and keeping it up-to-date so as to enable the next round of trajectory planning (Step 4).

\section{Use Case: Reconfiguration of a Robotic Arm}

For the evaluation of our approach, a DT has been developed to represent a robotic arm. With reference to the real physical environment a physical industrial demonstrator is used, comprising of a Niryo Ned2 robotics arm and industrial operational miniatures of real factory machinery controlled via real PLCs.

Each physical machine as well as the robot arm have corresponding virtual counterparts in the DT. The twin is constructed within the Unity game engine, and the instantiation of machines in the digital space is driven by the use of AutomationML (Automation Markup Language), which serves as the system's aforementioned DSL. When an AutomationML file is provided to Unity, it contains spatial and dimensional specifications for the machines. These are automatically translated into correctly dimensioned digital representations within the 3D simulation environment as described in the previous section. The installation can run both the physical robot and its Digital Twin simultaneously, allowing real-time data exchange and adaptive coordination between them. Real-time synchronization was made possible using ROS Noetic.  This setup allows the Unity-based DT to follow the physical robot's actions in real time. 

\begin{figure}[ht]
    \includegraphics[width=\columnwidth]{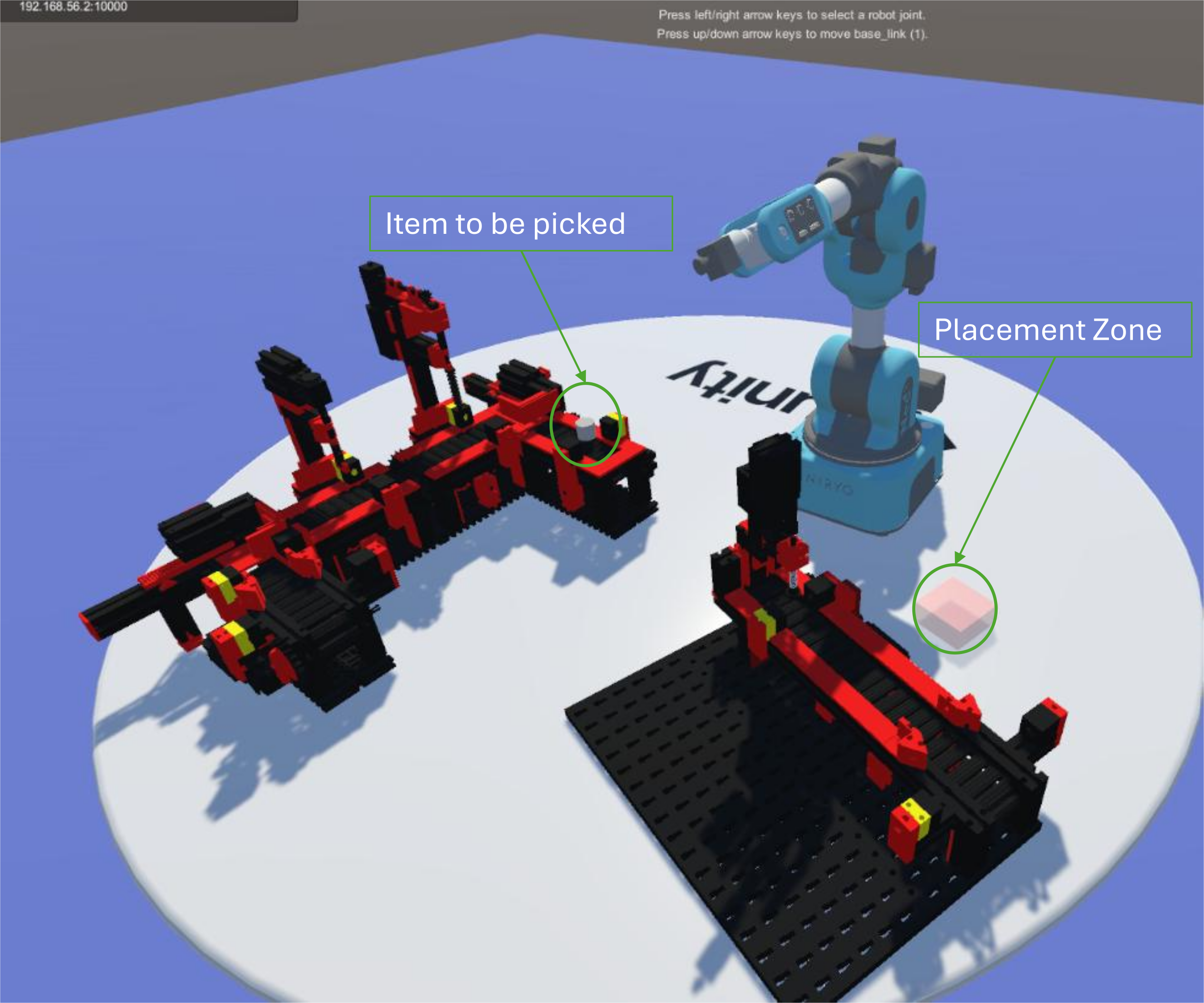}
    \caption{Scene topology}
    \label{fig:starting scene}
\end{figure}

\begin{figure}[ht]
    \includegraphics[width=\columnwidth]{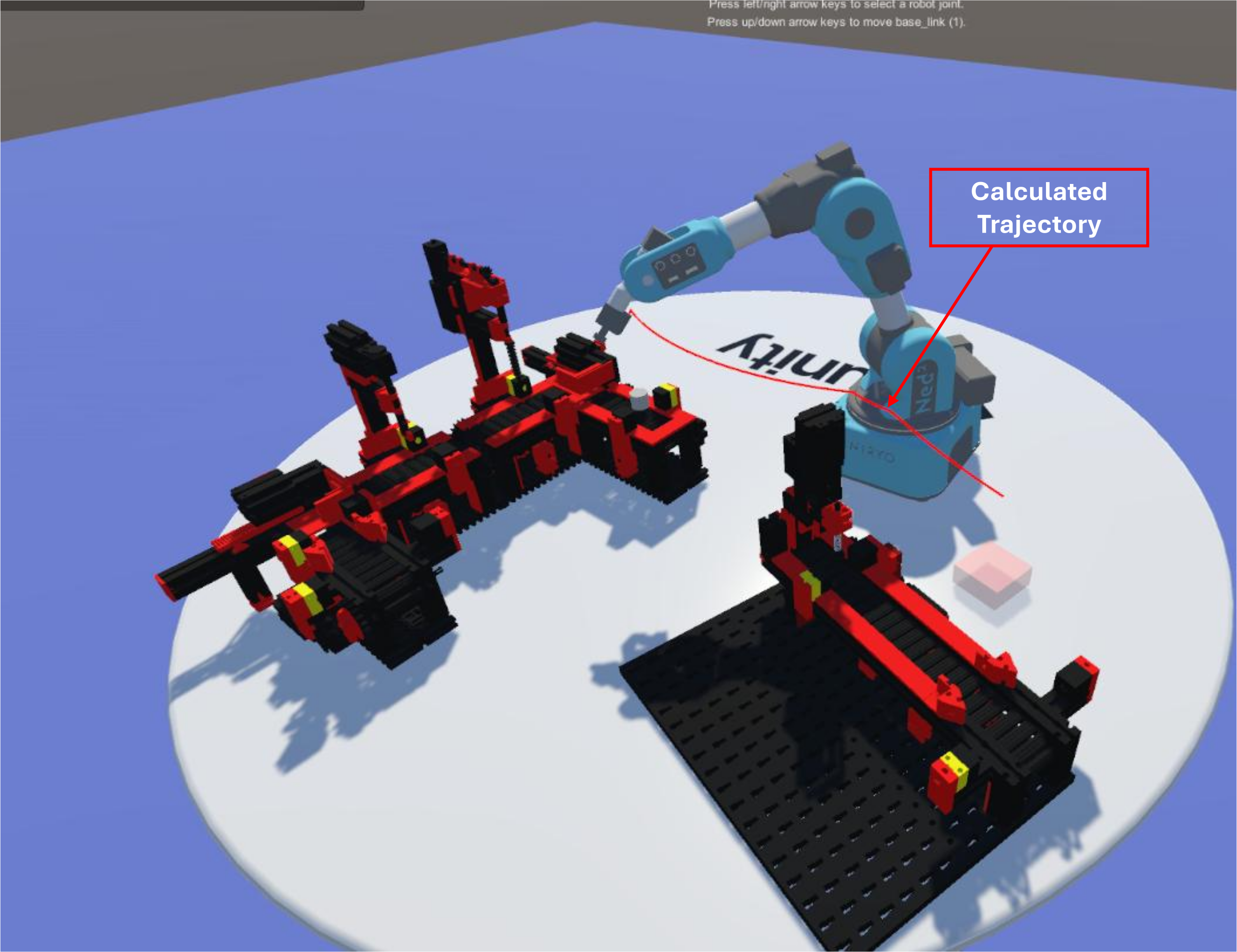}
    \caption{Visualized Trajectory}
    \label{fig:trajectory}
\end{figure}

This usecase scenario demonstrates what will happen if there is a change in the positions of the machines near the robot. The robot performs a pick-and-place task, picking a package from the machine, and placing it in a zone denoted as the "Target Placement" zone (Fig. \ref{fig:starting scene}). According to the AutomationML-instantiated DT, the 3D space is populated with machinery, therefore, the trajectory that the robot will follow to complete its task is non-trivial. As described in the previous sections, MoveIt! is called upon to calculate the trajectory optimally, and then visualize it in the DT. The visualization can be seen in Fig. \ref{fig:trajectory}. Once the trajectory is approved by the human operator, the real robot executes it in the real world, and the DT is updated synchronously. With the new conditions, the DT can make newly informed decisions and the reconfiguration cycle begins anew. 

\section{Conclusion}

The proposed DT-driven reconfiguration framework represents a significant advancement in robotic system adaptability, demonstrating how the strategic integration of widely accessible tools like Unity and ROS can substantially enhance robotic capabilities. By utilizing Unity as the spatial foundation for the Digital Twin, this approach unlocks an unprecedented range of applications that extend far beyond traditional robotic control systems. Unity's versatile platform enables not only high-fidelity environment simulation but also facilitates advanced functionalities such as real-time physics modeling, dynamic crowd simulation, and immersive VR/AR interfaces for human-robot interaction. The engine's integrated animation tools and Blender-like editors further enable sophisticated robot-environment interplay, supporting critical industrial features like animated machinery components and deformable meshes - capabilities that are particularly valuable in manufacturing and logistics applications.

The framework's effectiveness was rigorously validated through deployment in an industrial use case involving a robotic arm system operating in a dynamic demonstrator workspace. This implementation showcased the system's robust ability to detect and respond to topological changes in real-time, seamlessly recalculating optimal trajectories, visualizing updated motion plans, and synchronizing these adjustments with the physical robot. The successful closed-loop integration between the DT's simulation environment and the physical robotic system underscores not only the technical feasibility of this approach but also its practical value in settings where environmental variability is a constant challenge.

Looking forward, this research opens several promising paths for further development. The next phase will focus on enhancing the autonomy of the reconfiguration cycle through the integration of AI-driven decision-making algorithms. By combining the predictive capabilities of machine learning with the simulation power of the DT environment, future iterations could enable fully autonomous system reconfiguration in response to complex environmental changes. Additionally, substantial research efforts will be directed toward evaluating the framework's scalability potential, particularly in large-scale, heterogeneous environments such as smart city infrastructures and fully automated manufacturing lines. These complex operational contexts present unique challenges in terms of system interoperability, computational efficiency, and real-time performance that will need to be addressed.

In conclusion, the proposed DT-based framework represents a substantial step forward in robotic system design, offering a robust solution for autonomous adaptation in variable environments. By effectively bridging the gap between simulation and physical execution, the approach not only enhances current robotic applications but also lays the foundation for future developments in intelligent, self-configuring robotic systems. The combination of accessible technologies with advanced control strategies presented here provides a scalable model for the next generation of industrial robotics, with potential applications extending to fields as diverse as urban infrastructure management, disaster response, and space exploration. Future work will continue to refine these capabilities while exploring new applications that can benefit from this innovative approach to robotic system design.

\bibliographystyle{IEEEtran}
\bibliography{sample}

\end{document}